\documentclass[10pt, conference, compsocconf]{llncs}

\usepackage{times}
\usepackage{multirow}
\usepackage{comment}
\usepackage{amssymb}
\usepackage{graphicx}
\newcommand{\argmax}{\mathop{\rm argmax}\limits}

\usepackage{lscape}
\usepackage[whole]{bxcjkjatype}

\begin{document}
\title{Dynamic Fusion: Attentional Language Model for Neural Machine Translation}

\author{Michiki Kurosawa\inst{1} \and
Mamoru Komachi\inst{1}}

\institute{Tokyo Metropolitan University\\
6-6 Asahigaoka, Hino, Tokyo 191-0065, Japan\\
\email{kurosawa-michiki@ed.tmu.ac.jp, komachi@tmu.ac.jp}}

\maketitle

\begin{abstract}
Neural Machine Translation (NMT) can be used to generate fluent output.
As such, language models have been investigated for incorporation with NMT.
In prior investigations, two models have been used: a translation model and a language model.
The translation model's predictions are weighted by the language model with a hand-crafted ratio in advance.
However, these approaches fail to adopt the language model weighting with regard to the translation history. 
In another line of approach, language model prediction is incorporated into the translation model by jointly considering source and target information.
However, this line of approach is limited because it largely ignores the adequacy of the translation output.

Accordingly, this work employs two mechanisms, the translation model and the language model, with an attentive architecture to the language model as an auxiliary element of the translation model.
Compared with previous work in English--Japanese machine translation using a language model, the experimental results obtained with the proposed Dynamic Fusion mechanism improve BLEU and Rank-based Intuitive Bilingual Evaluation Scores (RIBES) scores. 
Additionally, in the analyses of the attention and predictivity of the language model, the Dynamic Fusion mechanism allows predictive language modeling that conforms to the appropriate grammatical structure.


\end{abstract}

\section{Introduction}
\label{section: intro}
With the introduction of deep neural networks to applications in machine translation, more fluent outputs have been achieved with neural machine translation (NMT) than with statistical machine translation \cite{modeling_coverage}.
However,  a fluent NMT output requires a large parallel corpus, which is difficult to prepare.
Therefore, several studies have attempted to improve fluency in NMT without the use of a large parallel corpus.

To overcome the data-acquisition bottleneck, the use of a monolingual corpus has been explored.
A monolingual corpus can be collected relatively easily, and has been known to contribute to improved statistical machine translation \cite{monolingual_SMT}.
Various attempts to employ a monolingual corpus have involved the following: pre-training of a translation model 
\cite{mono_pre}, initialization of distributed word representation \cite{hirasawa,pretrain_wordembedding}, and construction of a pseudo-parallel corpus by back-translation \cite{monolingual_backtrans}.

Here, we focus on a language modeling approach \cite{shallow_fusion,simple_fusion}.
Although recent efforts in NMT tend to output fluent sentences, it is difficult to reflect the linguistic properties of the target language, as only the source information is taken into consideration when performing translation \cite{linguistic_input_features}.
Additionally, language models are useful in that they contain target information that results in fluent output and can make predictions even if they do not know the source sentence.
In previous works utilizing a language model for NMT, both the language model and the conventional translation model have been prepared, wherein the final translation is performed by weighting both models.
In the Shallow Fusion mechanism \cite{shallow_fusion}, the output of the translation and language models are weighted at a fixed ratio.
In the Cold Fusion mechanism \cite{cold_fusion}, a gate function is created to dynamically determine the weight of the language model considering the translation model.
In the Simple Fusion mechanism \cite{simple_fusion}, outputs of both models are treated equally, whereas normalization steps vary.

In this research, we propose a ``Dynamic Fusion" mechanism that predicts output words by attending to the language model.
We hypothesize that each model should make predictions according to only the information available to the model itself; the information available to the translation model should not be referenced before prediction.
In the proposed mechanism, a translation model is fused with a language model through the incorporation of word-prediction probability according to the attention.
However, the models retain predictions independent of one another. 
Based on the weight of the attention, we analyze the predictivity of the language model and its influence on translation.

The main contributions of this paper are as follows:
\begin{itemize}
	\item We propose an attentional language model that effectively introduces a language model to NMT.
	\item We show that fluent and adequate output can be achieved with a language model in English--Japanese translation.
	\item We show that Dynamic Fusion significantly improves translation accuracy in a realistic setting.
	\item Dynamic Fusion's ability to improve translation is analyzed with respect to the weight of the attention.
\end{itemize}

\section{Previous works}
\subsection{Shallow Fusion}
Gulcehre et al. \cite{shallow_fusion} proposed Shallow Fusion, which translates a source sentence according to the predictions of both a translation model and a language model.
In this mechanism, a monolingual corpus is used to learn the language model in advance.
The translation model is improved through the introduction of the knowledge of the target language.

In Shallow Fusion, a target word $\hat{y}$ is predicted as follows:
\begin{equation}
\label{eq: shallow}
	\hat{y} = \argmax_{y} \log P_{{\rm TM}}({\bf y}|{\bf x}) + \lambda \log P_{{\rm LM}}({\bf y})
\end{equation}
where $\bf x$ is an input of the source language, $P_{{\rm TM}}({\bf y}|{\bf x})$ is the word-prediction probability according to the translation model, and $P_{\rm LM}({\bf y})$ is the word prediction probability according to the language model.
Here, $\lambda$ is a manually-determined hyper-parameter that determines the rate at which the language model is considered.

\subsection{Cold Fusion}
In addition to Shallow Fusion, Gulcehre et al. \cite{shallow_fusion} proposed Deep Fusion as a mechanism that could simultaneously learn a translation model and a language model.
Sriram et al. \cite{cold_fusion} extended Deep Fusion to Cold Fusion to pass information on a translation model for the prediction of a language model.

In this mechanism, a gating function is introduced that dynamically determines the weight, taking into consideration both a translation model and a language model.
Therein, the language model predicts target words by using information from the translation model.
Accuracy and fluency are improved through the joint learning of both models.

In Cold Fusion, a target word $\hat{y}$ is predicted as follows:
\begin{eqnarray}
\label{eq: cold}
	h_{{\rm LM}} &=& W_{{\rm LM}}S_{{\rm LM}}({\bf y})\\
	g &=& W_{{\rm gate}}[S_{{\rm TM}}({\bf y}|{\bf x}); h_{{\rm LM}}]\\
	h^{\prime} &=& [S_{{\rm TM}}({\bf y}|{\bf x}); g \cdot h_{{\rm LM}}]\\
	S_{\rm cold} &=& W_{{\rm output}}h^{\prime}\\
	\hat{y} &=& \argmax_{y} {\rm softmax}(S_{\rm cold})
\end{eqnarray}
where both $S_{{\rm TM}}({\bf y}|{\bf x})$ and $S_{\rm LM}({\bf y})$ are word-prediction logits\footnote{A logit is a probability projection layer without softmax.} with the translation model and the language model, respectively; $g$ is a function that determines the rate at which the language model is considered; $W_{\rm LM}$ ($|h|\times |V|$), $W_{\rm gate}$ ($2|h| \times |h|$), and $W_{\rm output}$ ($2|h| \times |V|$) are the weights of the neural networks; and $[a; b]$ is the concatenation of vectors $a$ and $b$.

\subsection{Simple Fusion}
Stahlberg et al. \cite{simple_fusion} proposed Simple Fusion, which simplifies Cold Fusion.
Unlike Cold Fusion, Simple Fusion does not use a translation model to predict words output by a language model.

For Simple Fusion, two similar methods were proposed: \textsc{PostNorm} (\ref{eq: postnorm}) and \textsc{PreNorm} (\ref{eq: prenorm}).
In \textsc{PostNorm} and \textsc{PreNorm}, a target word $\hat{y}$ is predicted as follows:
\begin{eqnarray}
\label{eq: postnorm}
	\hat{y} &=& \argmax_{y} {\rm softmax}({\rm softmax}(S_{{\rm TM}}({\bf y}|{\bf x})) \cdot P_{{\rm LM}}({\bf y}))\\
	\label{eq: prenorm}
	\hat{y} &=& \argmax_{y} {\rm softmax}(S_{{\rm TM}}({\bf y}|{\bf x}) + \log P_{{\rm LM}}({\bf y}))
\end{eqnarray}
where $S_{{\rm TM}}({\bf y}|{\bf x})$ denotes the word prediction logits with the translation model and $P_{{\rm LM}}({\bf y})$ denotes the word prediction probability according to the language model.

In \textsc{PostNorm}, the output probability of the language model is multiplied by the output probability of the translation model, wherein both models are treated according to the same scale.

In \textsc{PreNorm}, the log probability of the language model and the unnormalized prediction of the translation model are summed, wherein the language and translation models are treated with different scales.

Though the Simple Fusion model is relatively simple, it achieves a higher BLEU score compared to other methods that utilize language models.

 \begin{figure}[t]
  \center
  \includegraphics[width=8cm]{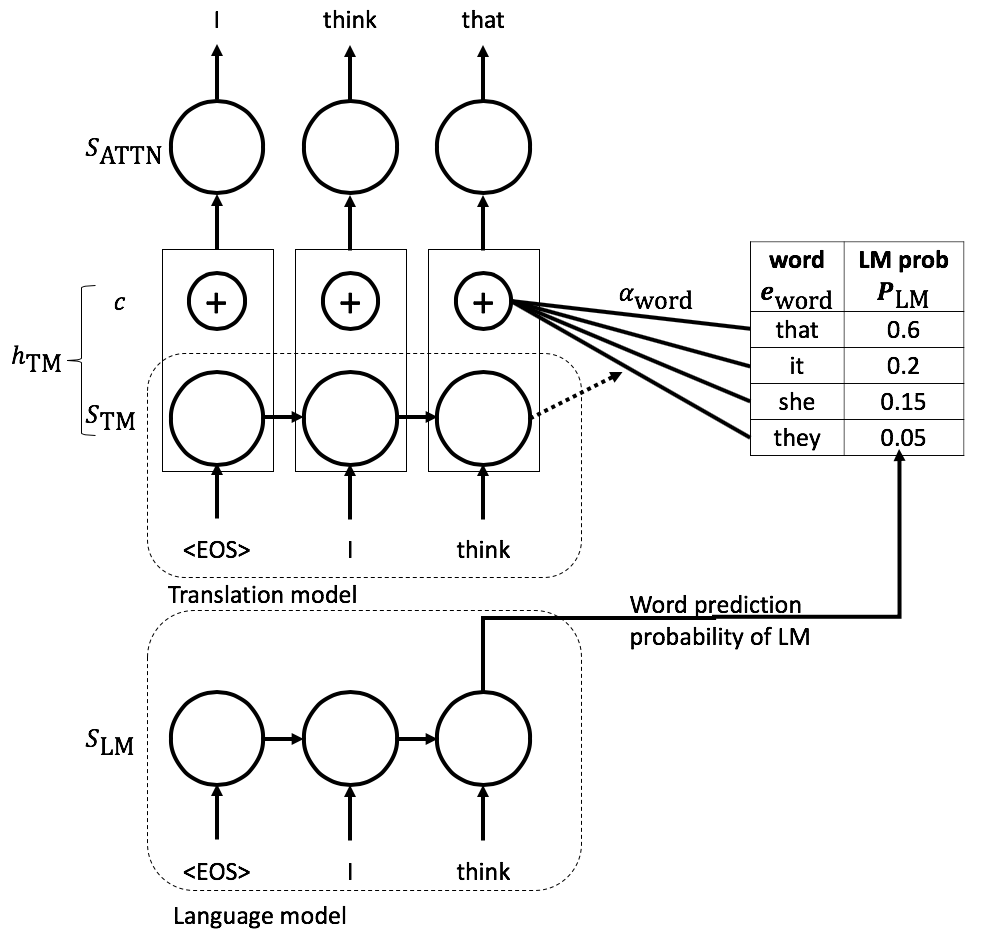}
  \caption{Dynamic Fusion mechanism.}
  \label{fig: attention}
\end{figure}

\section{Dynamic Fusion}
An attentional language model called ``Dynamic Fusion," is proposed in this paper.
In the Shallow Fusion and Simple Fusion mechanisms, information from the language model is considered with fixed weights.
However, translation requires that source information be retained, such that the consideration ratios should be adjusted from token to token.
Thus, both models should not be mixed with fixed weights.
The Cold Fusion mechanism dynamically determines the weights of mix-in; however, the Cold Fusion mechanism passes information from the translation model to the language model before prediction, and the language model thus does not make its own prediction.

Furthermore, in the previous research, it was necessary to make the vocabularies of the translation model and language model identical because the final softmax operation is performed in the word vocabulary dimension.
However, since the proposed mechanism mixes a language model as an attention, the vocabularies of the translation model and language model do not have to be completely consistent, and different word-segmentation strategies and subword units can be used.
Therefore, the proposed mechanism allows the use of a language model prepared in advance.

In the proposed mechanism, the language model serves as auxiliary information for prediction.
Thus, the language model is utilized independently of the translation model.
Unlike Cold Fusion, this method uses a language model's prediction score multiplied by word attention.

First, the word-prediction probability of the language model $P_{\rm LM}(y)$ is represented as follows:
\begin{equation}
	\label{eq: lm_dis_atte}
	P_{{\rm LM}}({\bf y}; y={\rm word}) = {\rm softmax}(S_{{\rm LM}}(\bf y))
\end{equation}
Next, hidden layers of the translation model attending to the language model $h_{\rm TM}$ are represented as follows:
\begin{eqnarray}
	\label{eq: h_atte}
	\alpha_{\rm word} &=& \frac{\exp(e_{\rm word}^{\rm T}S_{\rm TM}({\bf y}|{\bf x}))}{\sum_{{\rm word} \in V}\exp(e_{\rm word}^{\rm T}S_{\rm TM}({\bf y}|{\bf x})))}\\
	c_{{\rm word}} &=& \alpha_{\rm word} e_{{\rm word}} \\
	\label{eq: LMattention}
	\label{eq: c_LM}
	c_{\mathrm{LM}} &=& \sum_{{\rm word}} c_{{\rm word}} \cdot P_{{\rm LM}}({\bf y}; y={\rm word})\\
	h_{{\rm TM}} &=& [S_{{\rm TM}}({\bf y}|{\bf x});c_{\mathrm{LM}}]\\
	S_{\rm ATTN} &=& Wh_{{\rm TM}}
\end{eqnarray}
where $e_{{\rm word}}$ is the embedding of a word, $c_{{\rm word}}$ is the conventional word attention for each word, $c_{\mathrm{LM}}$ is the word attention's hidden state for the proposed Dynamic Fusion, and
$W$ ($2|h| \times {V}$) is a weight matrix of neural networks.
In Equation (\ref{eq: LMattention}), $c_{\mathrm{LM}}$ considers the language model by multiplying $P_{\rm LM}({\bf y}; y={\rm word})$ with a word attention.
In this mechanism, the prediction of the language model only has access to the target information up to the word currently being predicted.
Additionally, the language model and translation model can be made independent by using the conventional attention mechanism.

Finally, a target word $\hat{y}$ is predicted as follows:
\begin{equation}
	\label{eq: y_atte}
	\hat{y} = \argmax_{y} {\rm softmax}(S_{\rm ATTN})
\end{equation}

A diagram of this mechanism is shown in Figure \ref{fig: attention}, wherein the language model is used for the translation mechanism by considering the attention obtained from both the translation model and language model.

The training procedure of the proposed mechanism follows that of Simple Fusion and is performed as follows:
\begin{enumerate}
	\item A language model is trained with a monolingual corpus.
	\item The translation model and word attention to the language model are learned by fixing the parameters of the language model.
\end{enumerate}

\begin{table}[h]
  \begin{minipage}[t]{.45\textwidth}
	\caption{Corpus details．}
	\label{table: corpus}
	\small
	\begin{tabular}{crc}
	\hline
	&\# sentences&\begin{tabular}{c}\# maximum\\token\end{tabular}\\
	\hline
	\hline
	\begin{tabular}{c}Language model \\(monolingual)\end{tabular}&1,909,981&60\\
	Train (parallel)&827,188&60\\
	Dev (parallel)&1,790&\\
	Test (parallel)&1,812&\\
	\hline
	\end{tabular}
  \end{minipage}
  \hfill
  \begin{minipage}[t]{.45\textwidth}
	\caption{Experimental setting.}
	\label{table: experiment_setting}
	\small
	\begin{tabular}{ccc}
	\hline
	&setting\\
	\hline
	\hline
	Pre training epoch&15 epoch\\
	Maximum training epoch&100 epoch\\
	Optimization&AdaGrad\\
	Training rate&0.01\\
	Embed size&512\\
	Hidden size&512\\
	Batch size&128\\
	Vocabulary size (w/o BPE)&30,000\\
	\# BPE operation&16,000\\
	\hline
	\end{tabular}
   \end{minipage}
\end{table}

\section{Experiment}
\label{seq: experiment}
Here, the conventional attentional NMT \cite{bahdanau2014neural,luong-pham-manning:2015:EMNLP} and Simple Fusion models (\textsc{PostNorm}, \textsc{PreNorm}) were prepared as baseline methods for comparison with the proposed Dynamic Fusion model.
We performed English-to-Japanese translation.
Using this, the translation performance of the proposed model was evaluated by taking the average of two runs with BLEU \cite{papineni2002bleu} and Rank-based Intuitive Bilingual Evaluation Score (RIBES) \cite{RIBES}.
In addition, a significant difference test was performed using Travatar \footnote{http://www.phontron.com/travatar/evaluation.html} with 10,000 bootstrap resampling.
We performed an additional experiment on Japanese-to-English translation.
The details of the setting are the same as in English-to-Japanese translation, except that we only conducted the experiment once and did not perform a statistical significance test.

The experiment uses two types of corpora: one for a translation model and the other for a language model.
Thus, training data of the Asian Scientific Paper Excerpt Corpus (ASPEC) \cite{ASPEC} are divided into two parts: a parallel corpus and a monolingual corpus.
The parallel corpus, for translation, is composed of one million sentences with a high confidence of sentence alignment from the training data.
The monolingual corpus, for language models, is composed of two million sentences from the target side of the training data that are not used in the parallel corpus.
Japanese sentences were tokenized by the morphological analyzer MeCab \footnote{https://github.com/taku910/mecab} (IPADic), and English sentences were preprocessed by Moses \footnote{http://www.statmt.org/moses/} (tokenizer, truecaser).
We used development and evaluation set on the official partitioning of ASPEC as summarized in Table \ref{table: corpus}\footnote{We exclude sentences whose number of tokens with more than 60 tokens in training.}.
Vocabulary is determined using only the parallel corpus.
For example, words existing only in the monolingual corpus are treated as unknown words at testing, even if they frequently appear in the monolingual corpus to train the language model.
Additionally, experiments have been conducted with and without Byte Pair Encoding (BPE) \cite{luong-EtAl:2015:ACL-IJCNLP}.
BPE was performed on the source side and target side separately.

The in-house implementation \cite{WAT2017TMU} of the NMT model proposed by Bahdanau et al. \cite{bahdanau2014neural} and Luong et al. \cite{luong-pham-manning:2015:EMNLP} is used as the baseline model; all the other methods were created based on this baseline.
For comparison, settings are unified in all experiments (Table \ref{table: experiment_setting}).
In the pre-training process, only the language model is learned; the baseline performs no pre-training, as it does not have access to the language model.

\begin{table*}[tp]
   \caption{Results of English-Japanese translation. (Average of 2 runs.)}
   \label{table: result}
   \centering
   \small
   \begin{tabular}{cccccc|cc}
    \hline
    \multirow{2}{*}{Vocabulary}&TM&\multicolumn{2}{c}{w/o BPE}&\multicolumn{2}{c}{w/ BPE}&\multicolumn{2}{|c}{w/ BPE}\\

    &LM&\multicolumn{2}{c}{w/o BPE}&\multicolumn{2}{c}{w/ BPE}&\multicolumn{2}{|c}{w/o BPE}\\
    \hline
    &&BLEU&RIBES&BLEU&RIBES&BLEU&RIBES\\
    \hline
    \hline
    \multicolumn{2}{c}{Baseline}&31.28&80.78&32.35&81.17&32.35&81.17\\
    \multicolumn{2}{c}{\textsc{PostNorm}}&31.01&80.77&32.43&80.97&N/A&N/A\\
    \multicolumn{2}{c}{\textsc{PreNorm}}&31.61&80.78&32.69&81.24&N/A&N/A\\
    \multicolumn{2}{c}{Dynamic Fusion}&\textbf{31.84*}&\textbf{81.13*}&\textbf{33.22*}&\textbf{81.54*}&\textbf{33.05*}&\textbf{81.40*}\\
    \hline
   \end{tabular}
  \end{table*}

\begin{table*}[tp]
   \caption{Results of Japanese--English translation. (Single run.)}
   \label{table: jaen result}
   \centering
   \small
   \begin{tabular}{cccccc|cc}
    \hline
    \multirow{2}{*}{Vocabulary}&TM&\multicolumn{2}{c}{w/o BPE}&\multicolumn{2}{c}{w/ BPE}&\multicolumn{2}{|c}{w/ BPE}\\

    &LM&\multicolumn{2}{c}{w/o BPE}&\multicolumn{2}{c}{w/ BPE}&\multicolumn{2}{|c}{w/o BPE}\\
    \hline
    &&BLEU&RIBES&BLEU&RIBES&BLEU&RIBES\\
    \hline
    \hline
    \multicolumn{2}{c}{Baseline}&22.55&73.53&22.64&73.45&22.64&\textbf{73.45}\\
    \multicolumn{2}{c}{\textsc{PostNorm}}&21.47&73.21&22.09&72.77&N/A&N/A\\
    \multicolumn{2}{c}{\textsc{PreNorm}}&22.17&73.60&22.80&73.51&N/A&N/A\\
    \multicolumn{2}{c}{Dynamic Fusion}&\textbf{22.81}&\textbf{73.70}&\textbf{23.41}&\textbf{73.92}&\textbf{22.97}&\textbf{73.45}\\
    \hline
   \end{tabular}
  \end{table*}

\section{Discussion}
\subsection{Quantitative analysis}
The BLEU and RIBES scores results are listed in Table \ref{table: result} (English--Japanese) and Table \ref{table: jaen result} (Japanese--English).
In both scores, we observed similar tendencies with and without BPE.
Compared with the baseline model and the Simple Fusion model, Dynamic Fusion yielded improved results in terms of BLEU and RIBES scores.
However, between the baseline model and Simple Fusion, \textsc{PreNorm} improved but \textsc{PostNorm} was equal or worse.
Compared with \textsc{PreNorm}, Dynamic Fusion has improved BLEU and RIBES scores.
Accordingly, the improvement of the proposed method is notable, and the use of attention yields better scores.

In the English--Japanese translation, it was also confirmed that BLEU and RIBES were improved by using a language model.
RIBES was improved for the translation with Dynamic Fusion, suggesting that the proposed approach outputs adequate sentences.

The proposed method has statistically significant differences (p $<$ 0.05) in BLEU and RIBES scores compared to the baseline.
There was no significant difference between baseline and Simple Fusion, as well as between Simple Fusion and the proposed method.

In addition, we conducted additional experiments in a more realistic setting.
We experimented with the translation model in which BPE was performed, whereas the language model was trained on a raw corpus without BPE\footnote{We did not perform an experiment with Simple Fusion because Simple Fusion requires the vocabularies of both the language model and translation model to be identical.}.
It was found that the translation scores were improved as compared to the baseline model with BPE.

\subsection{Qualitative analysis}
Examples of the output of each model are giiven in Tables \ref{table:example} and \ref{table:example2}. 

In Table \ref{table:example}, compared with the baseline, the fluency of \textsc{PreNorm} and Dynamic Fusion resulted in improved translation.
Additionally, it can be seen that the attentional language model provides a more natural translation of the inanimate subject in the source sentence.
Unlike in English, inanimate subjects are not often used in Japanese.
Thus, literal translations of an inanimate subject sounds unnatural to native Japanese speakers.
However, \textsc{PostNorm} translates ``線量 (dose)" into ``用量 (capacity)”, which reduces adequacy.

\textsc{PreNorm} in Table \ref{table:example2} appears as a plain and fluent output.
However, neither of the Simple Fusion models can correctly translate the source sentence in comparison with the baseline.
In contrast, with Dynamic Fusion, the content of the source sentence is translated more accurately than in the reference translation; thus, without loss of adequacy, Dynamic Fusion maintains the same level of fluency.

This shows that the use of a language model contributes to the improvement of output fluency.
Additionally, Dynamic Fusion maintains relatively superior adequacy.

In Japanese--English translation, not only our proposed method but also other language models can cope with voice changes and inversion such as in Table \ref{table: jaen_good_ex}.
The use of active voice in Japanese where its counterpart is using passive voice is a common way of writing in Japanese papers \cite{yamagishi-etal-2017-improving}, and this example shows an improvement using a language model.

\subsection{Influence of language model}
Table \ref{table:typemiss} shows an example wherein the language model compensates for the adequacy.
In general, if there is a spelling error exists in the source sentence, a proper translation may not be performed owing to the unknown word.
In this example, the word ``temperature" is misspelled as ``temperture."
Thus, the baseline model translates the relevant part but ignores the misspelled word.
However, \textsc{PreNorm} and Dynamic Fusion complemented the corresponding part appropriately thanks to the language model.
The proposed method was able to translate without losing adequacy.
This result is attributed to the language model's ability to predict a fluent sentence.



\begin{landscape}
\begin{table*}[t]
   \caption{Example of fluency improvement by language model.}
   \label{table:example}
   \centering
   \small
   \scalebox{1}{
   \begin{tabular}{cc}
    \hline
    Model&Sentence (Output)\\
    \hline
    \hline
    Source&responding to these changes DERS can compute new dose rate .\\
    Reference&DERS は これら の 変化 に 対応 し て 新た な 線量 率 を 計算 できる 。\\
    \hline
    Baseline&これら の 変化 に 対応 する 応答 は , 新しい 線量 率 を 計算 できる 。\\
    &(Responses corresponding to these changes can calculate new dose rates.)\\
    Simple Fusion (\textsc{PostNorm})&これら の 変化 に 対応 する 応答 は 新しい 用量 率 を 計算 できる 。\\
    &(Responses corresponding to these changes can calculate new capacity rates.)\\
    Simple Fusion (\textsc{PreNorm})&これら の 変化 に 対応 する と , 新しい 線量 率 を 計算 できる 。\\
    &(In response to these changes, new dose rates can be calculated.)\\
    Dynamic Fusion&これら の 変化 に 対応 する こと により , 新しい 線量 率 を 計算 できる 。\\
    &(By responding to these changes, new dose rates can be calculated.)\\
    \hline
   \end{tabular}
   }
\end{table*}

\begin{table*}[t]
   \caption{Example of adequacy decline in Simple Fusion.}
   \label{table:example2}
   \centering
   \small
   \scalebox{1}{
   \begin{tabular}{cc}
    \hline
    Model&Sentence (Output)\\
    \hline
    \hline
    Source&\begin{tabular}{c}the magnetic field is given in the direction of a right angle or a parallel ( reverse to the flow ) to the tube axis . \end{tabular}\\
    Reference&磁場 は 管 軸 に 直角 か 平行 逆 方向 に 加え た 。 \\
    \hline
    Baseline&磁場 は 右 角 または 平行 ( 流れ ) の 方向 に 与え られ , 管 軸 に 平行 で ある 。\\
    &\begin{tabular}{c}(The magnetic field is given in the right angle or parallel (flow) direction and parallel to the tube axis.)\end{tabular}\\
    Simple Fusion (\textsc{PostNorm})&磁場 は 右 角度 または 平行 ( 流れ に 逆 に 逆 ) 方向 に 与え られ た 。\\
    &\begin{tabular}{c}(The magnetic field was applied at right angle or parallel (opposite to opposite to the flow) direction.)\end{tabular}\\
    Simple Fusion (\textsc{PreNorm})&磁場 は 右 角 または 平行 ( 流れ に 逆 方向 ) の 方向 に 与え られ た 。\\
    &\begin{tabular}{c}(The magnetic field was applied in the right angle or parallel (opposite to the flow) direction.)\end{tabular}\\
    Dynamic Fusion&磁場 は , 管 軸 に 直角 または 平行 ( 流れ に 逆 方向 ) の 方向 に 与え られる 。\\
    &\begin{tabular}{c}(The magnetic field is given in a direction perpendicular or parallel (reverse to the flow) to the tube axis.)\end{tabular}\\
    \hline
   \end{tabular}
   }
\end{table*}
\end{landscape}

\begin{table*}[t]
   \caption{Examples robust to changes in state.}
   \label{table: jaen_good_ex}
   \centering
   \small
   \scalebox{1}{
   \begin{tabular}{cc}
    \hline
    Model&Sentence (Output)\\
    \hline
    \hline
    Source&\begin{tabular}{c}変形 が 対 密度 分布 に 影響 し て いる こと が 分かっ た 。  \end{tabular}\\
    Reference&\begin{tabular}{c}it was found that the deformation gave effects to \\the pairing density distribution .\end{tabular}\\
    \hline
    Baseline&\begin{tabular}{c}it was found that deformation was affected by the pair density distribution .\end{tabular}\\
    \begin{tabular}{c}Simple Fusion \\(\textsc{PostNorm})\end{tabular}&\begin{tabular}{c}it was found that deformation affects the logarithmic density distribution .\end{tabular}\\
    \begin{tabular}{c}Simple Fusion \\(\textsc{PreNorm})\end{tabular}&\begin{tabular}{c}it was found that deformation affected the pair density distribution .\end{tabular}\\
    Dynamic Fusion&\begin{tabular}{c}it was found that the deformation affected the pair density distribution .\end{tabular}\\
    \hline
   \end{tabular}
   }
\end{table*}

\subsection{Influence of Dynamic Fusion}
\subsubsection{Fluency}
Excerpts from the output of Dynamic Fusion and word attention (top 5 words) are presented in Table \ref{table: attention}.

Except for the first token\footnote{The language model cannot predict that the first token correctly because it starts with \texttt{<BOS>}. }, the word attention includes the most likely outputs.
For example, if ``start bracket (「) " is present in the sentence, there is a tendency to try to close it with ``end bracket (」)".
Additionally, it is not desirable to close brackets with ``発電 (power generation)"; therefore, it predicts that the subsequent word is ``所 (plant)".
This indicates that the attentional language model can improve fluency while maintaining the source information.

Regarding attention weights, there are cases in which only certain words have highly skewed attention weights, among other cases in which multiple words have uniform attention weights.
The latter occurs when there are many translation options, such as the generation of function words on the target side. This topic requires further investigation.

\subsubsection{Adequacy} 
In contrast, it is extremely rare for Dynamic Fusion itself to return an adequate translation at the expense of fluency.
Even if a particular word has a significantly higher weight than other words, the prediction of the translation model may likely be used for the output if it changes the meaning of the source sentence.
In fact, the example in Table \ref{table: attention} contains many tokens in which the output of the language model is not considered, including at the beginning of the sentence.

One of the reasons for this is considered to be the difference in contributions between the translation model and the language model.
We decomposed the transformation weight matrix in Equation (\ref{eq: c_LM}) into the translation model and the language model matrices, and we calculated the Frobenius norm for each matrix.
The result reveals that the translation model contributes about twice as much as the language model.

\begin{landscape}
\begin{table*}[t]
   \caption{Comparison of adequacy by language model.}
   \label{table:typemiss}
   \centering
   \small
   \scalebox{0.95}{
   \begin{tabular}{cc}
    \hline
    Model&Sentence (Output)\\
    \hline
    \hline
    Source&\begin{tabular}{c}this paper explains the application of chemical processes utilizing supercritical phase\\where a liquid does not make phase change irrespective of temperture or pressure .  \end{tabular}\\
    Reference&\begin{tabular}{c}流体 が 温度 ・ 圧力 に かかわら ず 相 変化 し ない 状態 で ある 超 臨界 相 を 利用 し た 化学 プロセス の 応用 について 解説 し た 。\end{tabular}\\
    \hline
    Baseline&\begin{tabular}{c}液体 が 相 変化 を 持た ない 超 臨界 相 を 利用 し た 化学 プロセス の 応用 について 解説 し た 。\end{tabular}\\
    &\begin{tabular}{c}(The application of chemical processes using supercritical phase in which the liquid has no phase change is described.)\end{tabular}\\
    \begin{tabular}{c}Simple Fusion\\\end{tabular}&\begin{tabular}{c}液体 が 相 変化 を 起こす こと なく , 圧力 や 圧力 に 関係 なく 相 変化 を 生じる 化学 プロセス の 適用 について 解説 し た 。\end{tabular}\\
    (\textsc{PostNorm})&\begin{tabular}{c}(The application of the chemical process which causes the phase change regardless \\of the pressure and the pressure without the liquid causing the phase change is described.)\end{tabular}\\
    \begin{tabular}{c}Simple Fusion\\\end{tabular}&\begin{tabular}{c}液体 が 相 変化 を 起こさ ない 超 臨界 相 を 利用 し た 化学 プロセス の 応用 について , 温度 や 圧力 に 関係 なく 解説 し た 。\end{tabular}\\
    (\textsc{PreNorm})&\begin{tabular}{c}(The application of chemical processes using supercritical phase \\in which liquid does not cause phase change is described regardless of temperature and pressure.)\end{tabular}\\
    Dynamic Fusion&\begin{tabular}{c}液体 が 温度 や 圧力 に 関係 なく 相 変化 を 起こさ ない 超 臨界 相 を 利用 し た 化学 プロセス の 応用 について 解説 し た 。\end{tabular}\\
    &\begin{tabular}{c}(We have described the application of chemical processes that use a supercritical phase \\in which the liquid does not undergo a phase change regardless of temperature and pressure.)\end{tabular}\\
    \hline
   \end{tabular}
   }
\end{table*}

\begin{table*}[t]
   \caption{Dynamic Fusion output and attention example (excerpt).}
   \label{table: attention}
   \centering
   \scalebox{1.0}{
   \begin{tabular}{ccc|cc|cc|cc|cc|cc|cc}
    \hline
    モデル&\multicolumn{14}{c}{出力}\\
    \hline
    \hline
    Source&\multicolumn{14}{c}{details of dose rate of " Fugen Power Plant " can be calculated by using \texttt{<unk>} software .}\\
    Reference&\multicolumn{14}{c}{\texttt{<unk>} ソフトウエア を 用い て 「 ふ げん 発電 所 」 の 線量 率 を 詳細 に 計算 できる 。}\\
    \hline
    Dynamic Fusion&\multicolumn{14}{c}{「 ふ げん 発電 所 」 の 線量 率 の 詳細 を , \texttt{<unk>} ソフトウェア を 用い て 計算 できる 。 }\\
    &\multicolumn{14}{c}{(The details of the dose rate of "Fugen power plant", can be calculated by using the \texttt{<unk>} software.)}\\
    \begin{tabular}{c}Dynamic Fusion\\（excerpt）\end{tabular}&\multicolumn{2}{c|}{「}&\multicolumn{2}{c|}{ふ (Fu)}&\multicolumn{2}{c|}{げん (gen)}&\multicolumn{2}{c|}{発電 (Power)}&\multicolumn{2}{c|}{所 (Plant)}&\multicolumn{2}{c|}{」}&\multicolumn{2}{c}{の (of)}\\
    \multirow{5}{*}{\begin{tabular}{c}Word attention\\（Top5 word）\\ and weights\end{tabular}}&本&9.9e-1&この&5.5e-1&」&9.9e-1&」&1.0&所&9.9e-1&」&1.0&について&7.7e-1\\
    &標記&8.7e-5&その&3.5e-1&ね&3.2e-6&号&2.7e-8&機&1.3e-4&発電&3.2e-12&の&1.7e-1\\
    &この&4.2e-5&日本&7.0e-2&げん&2.0e-9&げん&1.4e-11&」&1.2e-6&の&1.7e-18&における&4.5e-2\\
    &また&8.5e-6&1&2.7e-2&出&1.1e-10&\texttt{<unk>}&1.1e-12&設備&7.7e-11&\texttt{<unk>}&7.6e-19&で&6.4e-3\\
    &これら&1.5e-6&高&4.7e-3&り&3.6e-11&・&1.8e-14&装置&2.6e-12&用&6.3e-19&と&3.2e-3\\
    \hline
   \end{tabular}}
  \end{table*}
\end{landscape}

\subsubsection{Role of language model}
Currently, most existing language models do not utilize the source information.
Accordingly, to eliminate noise in the language model's fluent prediction, language models should make predictions independently of translation models and thus be used in tandem with attention from translation models.
However, language models are useful in that they have target information that results in fluent output; they can thus make a prediction even if they do not know the source sentence.

Ultimately, the role of the language model in the proposed mechanism is to augment the target information in order for the translation model to improve the fluency of the output sentence.
Consequently, the fusion mechanism takes translation options from the language model only when it improves fluency and does not harm adequacy. It can be regarded as a regularization method to help disambiguate stylistic subtleness such as in the successful example in Table \ref{table:example}.

\section{Conclusion}
We proposed Dynamic Fusion for machine translation.
For NMT, experimental results demonstrated the necessity of using an attention mechanism in conjunction with a language model.
Rather than combining the language model and translation model with a fixed weight, an attention mechanism was utilized with the language model to improve fluency without reducing adequacy.
This further improved the BLEU scores and RIBES.


The proposed mechanism fuses the existing language and translation models by utilizing an attention mechanism at a static ratio.
In the future, we would like to consider a mechanism that can dynamically weight the mix-in ratio, as in Cold Fusion.




%

\bibliographystyle{splncs04}
\bibliography{reference}

\end{document}